\documentclass[pdflatex,sn-mathphys-num]{sn-jnl}

\usepackage{amsmath,amssymb,amsfonts}
\usepackage{amsthm}
\usepackage{mathrsfs}

\usepackage{graphicx}
\usepackage{wrapfig}
\usepackage{subfigure} 
\usepackage{svg}

\usepackage{array}
\usepackage{booktabs}
\usepackage{multirow}

\usepackage{algorithm}
\usepackage{algorithmicx}
\usepackage{algpseudocode}

\usepackage{textcomp}
\usepackage{mathptmx}  
\usepackage[utf8]{inputenc}

\usepackage{caption}
\usepackage{geometry}

\usepackage{tikz}
\usepackage{fontawesome5} 

\usepackage{manyfoot}
\usepackage{appendix}
\usepackage{lipsum}
\usepackage{ulem} 

\definecolor{lightgray}{gray}{0.95}        
\definecolor{headerblue}{RGB}{220,230,250} 
\definecolor{systemblue}{RGB}{70,130,200}
\definecolor{usergreen}{RGB}{34,139,34}
\definecolor{outputgray}{RGB}{95,95,95}
\definecolor{lightblue}{RGB}{240,248,255}
\definecolor{lightgreen}{RGB}{240,255,240}
\definecolor{lightgray}{RGB}{248,248,248}  
\definecolor{userblue}{RGB}{70,130,200}
\definecolor{originalgreen}{RGB}{34,139,34}
\definecolor{prunedorange}{RGB}{255,140,0}
\definecolor{bestpurple}{RGB}{138,43,226}
\definecolor{lightblue}{RGB}{240,248,255}
\definecolor{lightgreen}{RGB}{240,255,240}
\definecolor{lightorange}{RGB}{255,248,240}
\definecolor{lightpurple}{RGB}{248,240,255}
\definecolor{imagegray}{RGB}{248,248,250}
\usepackage{array}
\usepackage{booktabs}
\usepackage{tikz}
\usepackage{fontawesome5}
\usepackage{graphicx}
\usepackage{multirow}
\usepackage{fontawesome5}
\usepackage{listings}

\definecolor{systemblue}{RGB}{70,130,200}
\definecolor{usergreen}{RGB}{34,139,34}
\definecolor{outputgray}{RGB}{95,95,95}
\definecolor{lightblue}{RGB}{240,248,255}
\definecolor{lightgreen}{RGB}{240,255,240}
\definecolor{lightgray}{RGB}{248,248,248}

\definecolor{userblue}{RGB}{59,130,246}
\definecolor{originalgreen}{RGB}{16,185,129}
\definecolor{prunedorange}{RGB}{245,158,11}
\definecolor{bestpurple}{RGB}{139,92,246}
\definecolor{lightblue}{RGB}{239,246,255}
\definecolor{lightgreen}{RGB}{236,253,245}
\definecolor{lightorange}{RGB}{255,251,235}
\definecolor{lightpurple}{RGB}{245,243,255}
\definecolor{imagegray}{RGB}{249,250,251}
\definecolor{borderlight}{RGB}{229,231,235}

\usepackage{hyperref}

\begin{document}

\title[Article Title]{Compression Strategies for Efficient Multimodal LLMs in Medical Contexts}

\author[]{Tanvir A. Khan\thanks{1906048@eee.buet.ac.bd}}
\author[]{Aranya Saha\thanks{1906010@eee.buet.ac.bd}}
\author[]{Ismam N. Swapnil\thanks{1906064@eee.buet.ac.bd}}
\author[]{Mohammad A. Haque\thanks{arifulhoque@eee.buet.ac.bd}}

\affil[]{Department of Electrical and Electronic Engineering, Bangladesh University of Engineering and Technology (BUET), Dhaka, Bangladesh}

\abstract{
Multimodal Large Language Models (MLLMs) hold huge potential for usage in the medical domain, but their computational costs necessitate efficient compression techniques. This paper evaluates the impact of structural pruning and activation-aware quantization on a fine-tuned LLAVA model for medical applications. We propose a novel layer selection method for pruning, analyze different quantization techniques, and assess the performance trade-offs in a prune-SFT-quantize pipeline. Our proposed method enables MLLMs with 7B parameters to run within 4 GB of VRAM, reducing memory usage by \(70\%\) while achieving \(4\%\) higher model performance compared to traditional pruning and quantization techniques in the same compression ratio.
}


\maketitle
\section{Introduction}\label{sec1}

We present a unified and efficient pipeline for deploying multimodal large language models (MLLMs) in domain-specific settings, with a focus on dermatological visual question answering (VQA). Real-time clinical workflows require low-latency models, yet the substantial memory and computational demands of LLMs and MLLMs hinder their deployment in resource-constrained edge or cloud environments.

To tackle these challenges, we introduce a compression pipeline that integrates structural pruning, supervised fine-tuning (SFT), and quantization for task specific purposes. Our approach begins with structured pruning to remove redundant parameters, reducing the model’s size while preserving essential functionality. We employ a novel layer pruning criterion for domain specific applications. Since pruning can lead to performance drops, we leverage SFT to restore task-specific capabilities. Following pruning and fine-tuning, we incorporate Activation-aware Weight Quantization, which significantly boosts memory efficiency with minimal performance degradation. We show that traditional calibration free quantization techniques (such as bitsandbytes \cite{dettmers2022llmint88bitmatrixmultiplication}, hqq \cite{badri2023hqq}) fails to retain language modeling capacity of pruned LLMs. 

We validate our method on dermatological VQA tasks using the LLaVA (Large Language and Vision Assistant) \cite{Liu2023VisualIT} model, demonstrating its effectiveness for skin disease diagnosis. Our results show that compressed MLLMs can retain strong performance while being suitable for deployment in real-world clinical environments.

\section{Related Work}\label{sec2}
\subsection{Multi-modal Large Language Models}
Large Language Models (LLMs) have demonstrated remarkable capabilities in understanding and generating human-like text across diverse domains. These models, trained on vast amounts of textual data, leverage deep learning architectures such as transformers to capture complex linguistic patterns. Despite their success in natural language processing (NLP), traditional LLMs are limited to text-based inputs, restricting their applicability in fields requiring multimodal understanding\cite{Wu2023MultimodalLL}. To address this limitation, \textbf{Multimodal Large Language Models (MLLMs)} extend LLMs by integrating multiple data modalities, such as images, audio, and structured medical data, alongside text. By incorporating vision-language pretraining techniques, these models can interpret and generate responses based on both textual and visual information, making them highly effective in perception-driven tasks. 
In healthcare, MLLMs can play a crucial role in medical image interpretation, diagnostic assistance, and patient-doctor interaction modeling\cite{Santomauro2023EnhancingMI}. They can analyze dermatological images, radiology scans, and pathology slides while cross-referencing clinical notes for comprehensive assessments. Additionally, MLLMs enhance medical chatbots and virtual assistants, improving accessibility to healthcare information and decision support for professionals. By bridging the gap between language and vision, these models contribute to more accurate and context-aware medical diagnostics. As MLLMs continue to evolve, they hold the potential to revolutionize healthcare by enabling AI-driven, multimodal clinical decision-making and patient care.
However, the deployment of \textbf{Multimodal Large Language Models (MLLMs)} is constrained by the high cost of model serving. LLMs contain billions of parameters, making their inference computationally expensive, particularly for consumer applications. Deploying these models on GPUs requires significant memory and processing power, which may not be feasible for many organizations. In the medical domain, where data privacy and security are paramount, cloud-based solutions are often impractical, necessitating \textbf{on-premise deployment}. To mitigate the computational burden and reduce serving costs, \textbf{model compression techniques} can be applied. Pruning and Quantization are the most popular model compression techniques. These techniques, when applied effectively, enable efficient deployment of MLLMs without substantial performance degradation. By leveraging compression, MLLMs can be made more accessible and cost-effective, facilitating their adoption in real-world healthcare applications. 

\subsection{Pruning}

Pruning removes less significant parameters from a model, reducing memory and computational requirements while preserving core functionality. However, pruning large language models (LLMs), particularly Transformer-based architectures, presents unique challenges due to their dense attention mechanisms and inter-layer dependencies.

Pruning can be broadly classified into two types, Unstructured and Structured. Structural pruning is preferred when it is important to save memory resources as it does not require specialized hardware optimized for sparse matrix multiplication. A notable development in structural pruning is \textit{depth pruning}, which involves removing entire Transformer blocks. Kim et al.~\cite{kim2024shortened} explored this technique in LLaMA, showing that selective removal of less impactful layers leads to large speedups in inference time. They further highlighted the importance of continued pretraining (CPT) to restore performance after aggressive pruning. Depth pruning presents a compelling solution for memory-constrained deployment, especially when coupled with fine-tuning strategies to recover performance.

Different works have proposed different algorithms to determine which layers of model contributed the least to the generation of output. The baseline method for layer importance detection is weight magnitude \cite{han2015learning}. The sum of weight magnitude is taken as the layer importance and the layers with the least magnitude are removed. ShortGPT \cite{men2024shortgptlayerslargelanguage} uses the cosine similarity between the embeddings of the input and output of a layer to derive layer importance.


\subsection{Quantization}

\textbf{Quantization} is a compression technique that maps floating-point values to low-bit integer representations, reducing both memory usage and computation cost. It can be expressed for a given weight group $w$ and an input $x$ as a typical linear transformation is $y = wx$. In standard post-training quantization, the quantized weights $\hat{w}$ are computed as:

\begin{equation}
    Q(w) = \Delta \cdot \text{Round}\left( \frac{w}{\Delta} \right), \quad \Delta = \frac{\max(|w|)}{2^{N-1}},
\end{equation}

where $N$ is the bit-width (e.g., 3 or 4), and $\Delta$ is the quantization scale derived from the maximum absolute value in the weight group.
Quantization is broadly categorized into Quantization-Aware Training (QAT) and Post-Training Quantization (PTQ).

QAT integrates quantization during training, using backpropagation to update quantized weights~\cite{liu2023llm}. While effective, it is computationally intensive and difficult to scale to LLMs. Recent works such as QLoRA, PEQA, and LoftQ~\cite{dettmers2023qlora} combine QAT with Parameter-Efficient Fine-Tuning (PEFT) to address this limitation. L4Q~\cite{jeon2024l4q} further introduces a LoRA-wise learned step size to improve generalization across tasks.

In contrast, PTQ offers a training-free alternative, making it highly suitable for LLM compression. Common configurations include weight-only INT4 (W4A16) and full INT8 (W8A8) quantization. PTQ can be subdivided into weight-only, weight-activation, and key-value (KV) cache quantization.

Among PTQ methods, Activation-aware Weight Quantization (AWQ)~\cite{lin2024awq} has gained prominence. AWQ identifies and preserves 0.1\%--1\% of high-impact weights by analyzing activation distributions instead of weight magnitudes. It applies per-channel scaling to protect important features, maintaining hardware efficiency while achieving strong performance in both vision and language tasks. AWQ outperforms traditional round-to-nearest (RTN) methods and rivals mixed-precision approaches without the hardware complexity.

There are few works that addresses both pruning and quantizaiton. 
Kim et al.~\cite{kim2023quantization} proposed Quantization Robust Pruning With Knowledge Distillation for convlutional neural networks. Xu et al.~\cite{xu2021accelerating} proposed structural pruning and quatization to accelerate Federated learning techniques.


\section{Methodology}\label{sec3}
\begin{figure}[h]
  \centering
  \includegraphics[width=\textwidth]{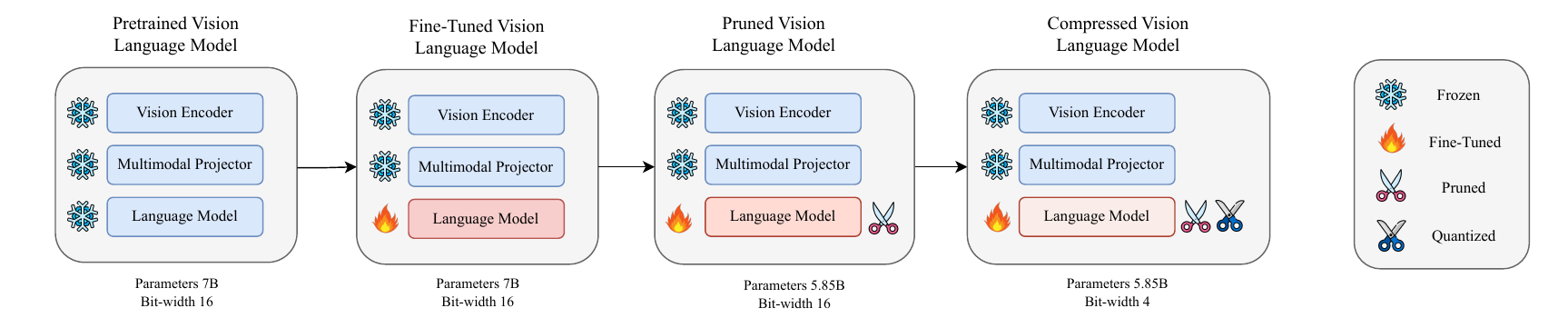}
  \caption{Overall pipeline of the proposed methodology.}
  \label{fig:overall_pipe}
\end{figure}

We apply pruning and quantization on a Multimodal large language model (LLaVA) to investigate the effect of compression techniques on the model. 
The overall pipeline of our methodology is shown in figure~\ref{fig:overall_pipe}.

\subsection{Dataset and Fine-Tuning}

For this study, we utilize the DermNet dataset, an open-source collection designed for dermatological image analysis. The dataset comprises approximately 19,500 images obtained from a diverse range of patients with various skin conditions. It provides diagnostic labels for 23 high-level disease categories, which are further subdivided into 642 fine-grained subcategories. These classes encompass a wide spectrum of dermatological conditions, including bacterial, viral, fungal, inflammatory, and malignant skin diseases.  

Given the scope of our research, we focus on a subset of eight disease categories: Rosacea, Actinic Keratosis, Basal Cell Carcinoma, Dermatitis, Melanoma, Psoriasis, Lichen Planus, and Seborrheic Keratoses. The dataset used for model training consists of 1,737 images, while the test set includes 39 images. This selection enables us to evaluate the model’s effectiveness in distinguishing between clinically relevant dermatological conditions.  
Conversation format:
\begin{itemize}
    \item \textbf{Human:} ``What is the name of this disease?''
    \item \textbf{Response:} ``This is seborrheic keratosis.''
    \item \textbf{Human:} ``What is a seborrheic keratosis?''
    \item \textbf{Response:} ``Seborrheic keratosis (SK) is a benign skin growth common in aging adults, typically warty, waxy, or scaly, and not linked to cancer.''
\end{itemize}
We apply supervised finetuning to the instruction tuning dataset using QLoRA with \(rank=8\) and \(alpha=32\). All of our subsequent experiments are performed on the fine-tuned version of LLaVA.

\subsection{Structural Pruning}

In structural pruning, the model is first analyzed to identify redundant components that have minimal impact on overall performance. We leverage a small calibration dataset to determine these non-critical parameters. To balance compression effectiveness with hardware efficiency, we constrain the granularity of pruning to entire Transformer layers. This coarse-grained approach enables practical acceleration on existing hardware while maintaining the model's core functionality.

The LLaVA model comprises three modules : the CLIP encoder, the multi-modal projector, and the language model. Among these, the language model contains over 20 times the parameters of the other two modules combined, making it the primary target for pruning.

Within the language model, there is a token embedding layer, 32 transformer layers, and a language model head. The token embedding layer and the language model head are essential for information encoding, so we focus on pruning the transformer blocks.

For structured pruning, we treat each block of the language model as an individual pruning unit similar to the method introduced by Kim et al.\cite{kim2024shortened} . Blocks 1, 2, and 32 are excluded from pruning, as their removal leads to significant degradation in the overall model performance.\cite{kim2024shortened}\cite{ma2023llm}


\begin{figure}[h]
    \centering
    \includegraphics[width=.9\linewidth]{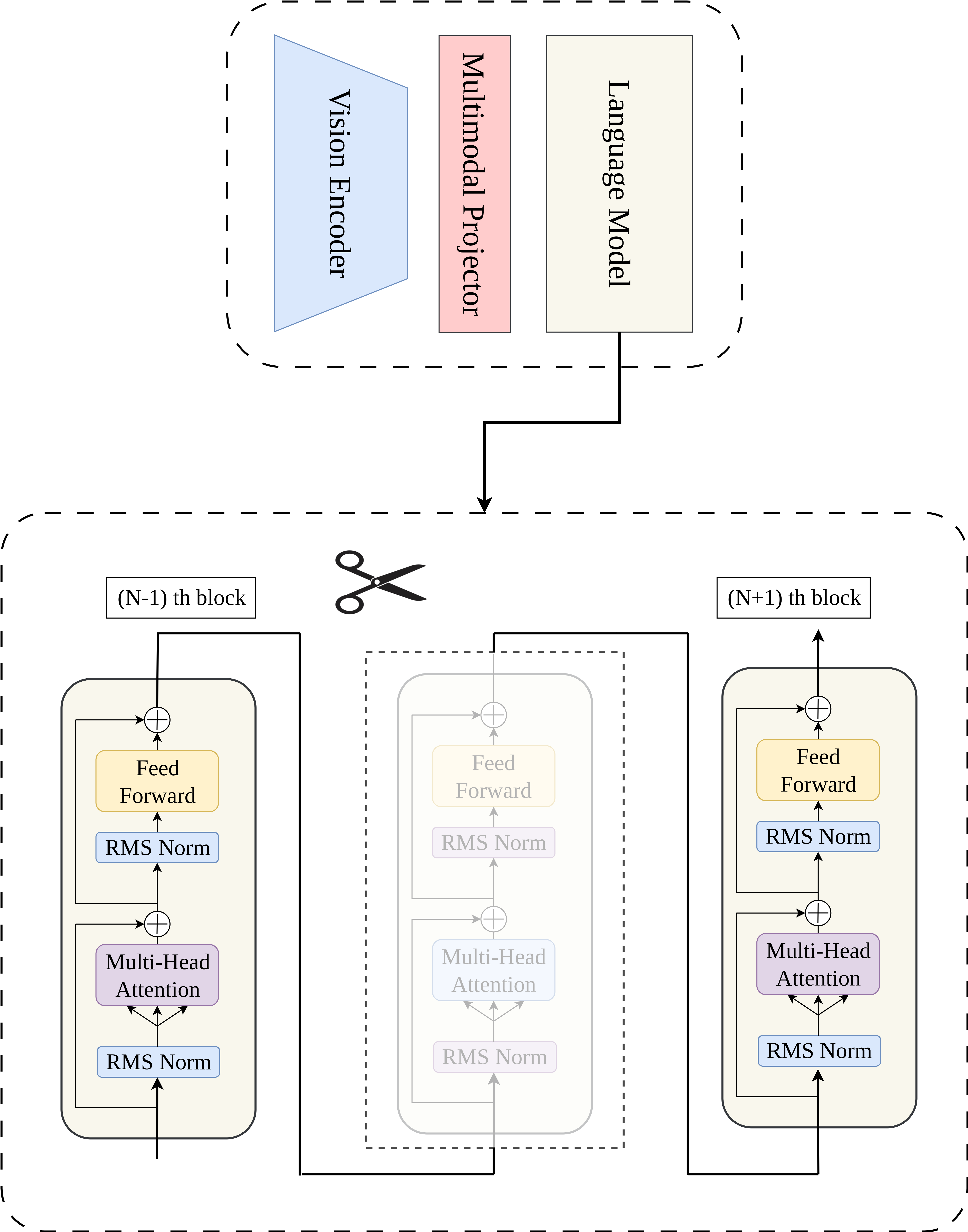}
    \caption{Workflow of LLM's Structured Pruning}
    \label{fig:llm_struct_prune}
\end{figure}

To demonstrate the effect of layer removal, we prune one block at a time from the language model of the LLaVA model as shown by figure~\ref{fig:llm_struct_prune}. Then we use a calibration dataset by randomly sampling from our question-answering dataset and run forward passes. 
For a given input sample $x_i \in \mathcal{D}_{\text{cal}}$, we compute the output representation from both the pruned model $f_{\text{pruned}}$ and the original model $f_{\text{orig}}$:

\begin{equation}
    y_i^{\text{pruned}} = f_{\text{pruned}}(x_i), \quad y_i^{\text{orig}} = f_{\text{orig}}(x_i)
\end{equation}

We store the outputs from the forward passes. We then compare the outputs using cosine similarity. To compute the cosine similarity we utilize sentence transformer \cite{wang2020minilmdeepselfattentiondistillation} embedding model  \(all-MiniLM-L6-v2\). We denote the sentence transformer model as  $g(\cdot)$.  The embeddings are calculated as \begin{equation}
    e_i^{\text{pruned}} = g(y_i^{\text{pruned}}), \quad e_i^{\text{orig}} = g(y_i^{\text{orig}})
\end{equation}

The cosine similarity between the two embeddings is then computed as:

\begin{equation}
    S_i = \frac{ e_i^{\text{pruned}} \cdot e_i^{\text{orig}} }{\| e_i^{\text{pruned}} \| \| e_i^{\text{orig}} \|}
\end{equation}

where $\| \cdot \|$ represents the Euclidean norm.

\begin{figure}[h]
    \centering
        \includegraphics[width=.7\textwidth]{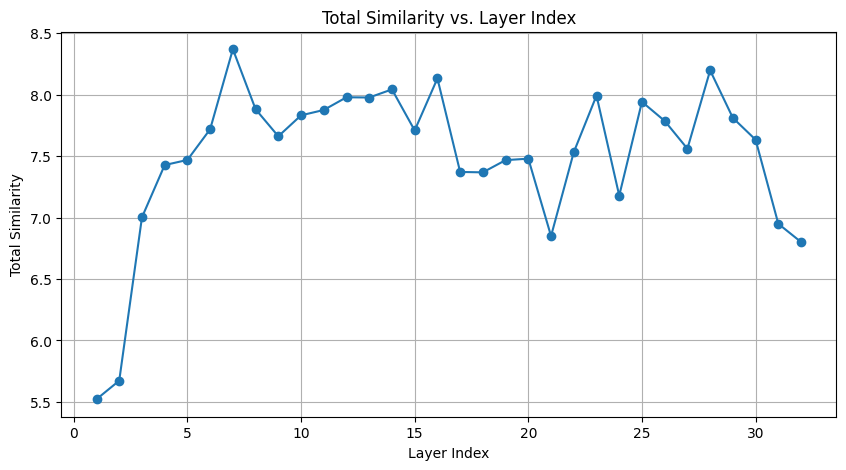}
        \caption{Similarity score across different layers obtained using Equation~\ref{eq:5}, illustrating the variance in contribution to overall model performance.}
        \label{fig:importance_score_layers}
\end{figure}

Finally, the average cosine similarity for a pruned block is determined by computing the average cosine similarity across all samples in the calibration dataset:

\begin{equation} \label{eq:5}
    S_{\text{avg}} = \frac{1}{|\mathcal{D}_{\text{cal}}|} \sum_{i=1}^{|\mathcal{D}_{\text{cal}}|} S_i
\end{equation}
The average cosine similarity obtained for each layer removed hints the significance of that layer for preserving the performance of the model. Figure~\ref{fig:importance_score_layers} shows the total similarity between the outputs of the original model and the outputs of the model after pruning a single transformer layer. A higher similarity indicates that the model can produce coherent output without the pruned layer, indicating lower layer importance. The less important layers are then pruned iteratively. Sample outputs of iterative pruning are shown in Appendix~\ref{secA2}

\subsection{Supervised Finetuning after Pruning}
After pruning, the performance of our LLaVA model remained stable when up to 6 transformer layers were removed. However, applying quantization to further compress the model led to a noticeable drop in accuracy. To mitigate this degradation, we employed Supervised Fine-Tuning (SFT) using the Dermnet dataset, which was also used during the initial training phase, prior to quantization. SFT fine-tunes the remaining model weights, enabling the model to maintain moderate performance despite structural reductions.

While Continuous Pretraining (CPT) on a large-scale corpus, as proposed by Kim et al.~\cite{kim2024shortened}, could potentially offer superior performance recovery, our computational constraints restricted us to SFT. Nonetheless, this approach allowed the pruned model to partially regain its effectiveness by adapting its remaining parameters to the target domain data.

\subsection{Post Training Quantization}  
To mitigate the loss of critical information during quantization, we adopt \textit{activation-aware quantization}, a strategy that selectively preserves salient weights based on activation patterns, as proposed by Lin et al.~\cite{lin2024awq}. Traditional quantization techniques often clip outlier activation values—especially within the self-attention layers of large language models—resulting in degraded performance due to the suppression of expressive, high-magnitude features.

\begin{figure}[h]
    \centering
    \includegraphics[width=0.75\linewidth]{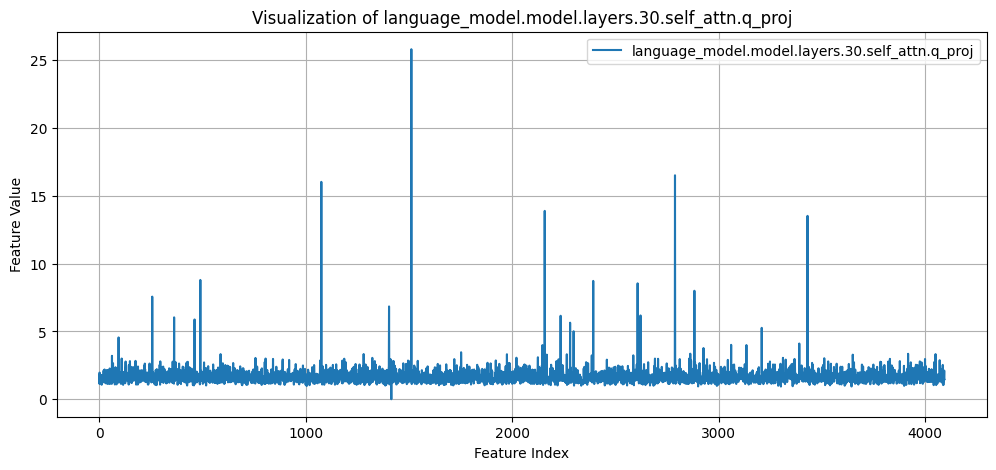}
    \caption{Distribution of activation values in a self-attention layer.}
    \label{fig:activation_dist}
\end{figure}

Figure~\ref{fig:activation_dist} illustrates the distribution of activation values at the output of the query projection (\texttt{q\_proj}) matrix in the 30th transformer layer of the language model. The input tensor is projected using a query weight matrix of shape \(4096 \times 4096\), resulting in an output activation matrix of shape \(\text{input\_size} \times 4096\). While the majority of activation values lie within the range of 0 to 3, several prominent outliers exceed a value of 10, with the highest peak surpassing 25. These outliers suggest the presence of sparsely distributed, high-magnitude activations, which may play a critical role in the model’s attention mechanism and information propagation.

To mitigate the quantization error of important weights while maintaining hardware efficiency, AWQ adopts a salient weight protection strategy by scaling up the weights of salient channels while inversely scaling the corresponding input activations.

 We modify the quantization pathway by scaling the weight $w$ with a factor $s > 1$ and inversely scaling the input $x$:

\begin{equation}
    Q(w \cdot s) \left( \frac{x}{s} \right) = \Delta' \cdot \text{Round}\left( \frac{ws}{\Delta'} \right) \cdot \frac{x}{s},
\end{equation}

where $\Delta'$ is the new quantization scale. Empirical evidence suggests that scaling a single weight element rarely changes the group's maximum, hence $\Delta' \approx \Delta$.

To determine the optimal scaling factor $s$, we minimize the quantization error using a small calibration set. The loss function is defined as:

\begin{equation}
    \mathcal{L}(s) = \left\| Q\left(W \cdot \text{diag}(s)\right) \cdot \left(\text{diag}(s)^{-1} \cdot X\right) - WX \right\|_2,
\end{equation}

where $W$ is the original weight matrix, $X$ is the activation matrix from the calibration set, and $s$ is a per-channel scaling vector.

To make the search efficient and stable, AWQ restricts the scaling factors based on the per-channel average activation magnitude $s_X$. The scaling vector $s$ is parameterized by an exponent $\alpha$:

\begin{equation}
    s = s_X^{\alpha}, \quad \alpha^* = \underset{\alpha}{\arg\min} \; \mathcal{L}(s_X^{\alpha}).
\end{equation}

The optimal $\alpha$ is then identified via a grid search.

\section{Experiments}\label{sec4}
This section details the experimental setup designed to evaluate our proposed compression pipeline. We assess the trade-offs between model performance and computational efficiency (VRAM usage, latency) at each stage. All experiments were conducted on Kaggle's T4 GPUs, and the source code is available on \href{https://github.com/takakib123/LLaVA_Prune}{Github}.

\subsection{Baselines}
We compare our method against the following baselines:
\begin{itemize}
    \item \textbf{Original LLaVA (FP16)}: Full-precision LLaVA without compression.
    \item \textbf{Isolated Pruning}: Structured pruning without any quantization.
    \item \textbf{Isolated Quantization}: Post-training quantization without pruning.
\end{itemize}

\subsection{Evaluation Metrics}
To assess model quality in the visual question answering task, we employ an LLM-based judge (detailed in Appendix \ref{secA2}) to generate a \textbf{performance score}. This approach was chosen over traditional metrics to better capture the nuance and clinical relevance of the generated text. For computational efficiency, we measure \textbf{peak VRAM usage} (GB) and \textbf{inference latency} ( ms/token ).

\subsection{Structural Pruning}
Structural pruning removes parameters from the model. While it reduces the number of parameters providing faster inference and less VRAM usage, it also degrades the model's conversation capability to some extent. After removing each block, we obtain an importance score for each layer in the model. A higher score indicates better output accuracy even without the corresponding layer, signifying the lower importance of that block. Then, we removed entire layers from the model based on the importance scores. Layers with minimum sum were identified and pruned iteratively. Layers with the lowest importance scores were then iteratively removed. The results, presented in Table~\ref{tab:structured_pruning_llm}, illustrate the trade-off between model performance and efficiency. As the table shows, performance remains relatively stable with up to four layers removed (a 65.75 score at 10\% compression), but degrades sharply thereafter. Removing ten layers (29\% compression) causes the performance score to collapse to 27.25, indicating that crucial model capabilities have been lost.

\begin{table}[h]
    \caption{Results of Structured Pruning by Removing Layers}
    \label{tab:structured_pruning_llm}
\begin{tabular}{ccccc}
\hline
\textbf{\begin{tabular}[c]{@{}c@{}}Compression\\ Ratio(\%)\end{tabular}} & \textbf{\begin{tabular}[c]{@{}c@{}}Number of\\ Parameters\end{tabular}} & \textbf{\begin{tabular}[c]{@{}c@{}}Number of\\ Layers Removed\end{tabular}} & \textbf{\begin{tabular}[c]{@{}c@{}}Performance\\ Score\end{tabular}} & \textbf{VRAM (GB)} \\ \hline
0                                                                        & 7.063B                                                                  & 0                                                                           & 80.12                                                                & 13.4               \\
5                                                                        & 6.659B                                                                  & 2                                                                           & 73.15                                                                & 12.5               \\
10                                                                       & 6.254B                                                                  & 4                                                                           & 65.75                                                                & 11.7               \\
20                                                                       & 5.647B                                                                  & 7                                                                           & 57.25                                                                & 10.5               \\
23                                                                       & 5.444B                                                                  & 8                                                                           & 49.50                                                                & 10.1               \\
26                                                                       & 5.242B                                                                  & 9                                                                           & 36.25                                                                & 9.7                \\
29                                                                       & 5.04B                                                                   & 10                                                                          & 27.25                                                                & 9.4                \\ \hline
\end{tabular}
\end{table}

\subsection{Supervised Finetuning}

To counteract the performance degradation caused by pruning, we applied supervised finetuning (SFT) to the pruned models using the Qlora method (rank = 16, alpha = 8). Table~\ref{tab:sft_result} demonstrates that SFT is highly effective at recovering performance. For example, the 5.6B parameter model's score improved from 62.50 to 74.25 after finetuning. The table also confirms that finetuning a smaller model is more efficient, requiring less VRAM and time.

\begin{table}[h]
    \caption{Results of Supervised Finetuning After Pruning}
    \label{tab:sft_result}
\begin{tabular}{lcccc}
\hline
\multicolumn{1}{c}{\textbf{Model}} & \textbf{\begin{tabular}[c]{@{}c@{}}Time for \\ Finetuning\end{tabular}} & \textbf{VRAM} & \textbf{\begin{tabular}[c]{@{}c@{}}Performance Score \\ Before Finetuning\end{tabular}} & \textbf{\begin{tabular}[c]{@{}c@{}}Performance Score\\ After Finetuning\end{tabular}} \\ \hline

\textbf{Pruned LLaVA (5.6 B)}         & 2 hour 42 min                                                           & 13.2          & 62.50                                                                                   & 74.25                                                                                 \\
\textbf{Pruned LLaVA (5.2 B)}       & 2 hour 14 min                                                           & 10.8          & 38.25                                                                                   & 52.25                                                                                 \\ \hline
\end{tabular}
\end{table}

\subsection{Quantization}

Table~\ref{tab:quant_result} shows the results of quantization after structured pruning and supervised finetuning. We show results for both AWQ and Bitsandbytes quantization technique. Quantization was performed with group size = 64 for both methods. Both method require same VRAM usage.

\begin{table}[h]
    \caption{Results of Quantization}
    \label{tab:quant_result}
\begin{tabular}{lccc}

\hline
\multicolumn{1}{c}{\multirow{2}{*}{\textbf{Model}}} & \multicolumn{2}{c}{\textbf{Performance Score}} & \multirow{2}{*}{\textbf{\begin{tabular}[c]{@{}c@{}}VRAM\\ (GB)\end{tabular}}} \\
\multicolumn{1}{c}{}                                & \textbf{AWQ}      & \textbf{Bitsandbytes}      &                                                                               \\ \hline
\textbf{Base LLaVA (7B)}                            & 64.72             & 66.25                      & 4.5                                                                           \\
\textbf{Pruned LLaVA (5.6B)}                        & 54.25             & 47.25                      & 3.9                                                                           \\
\textbf{Pruned LLaVA (5.2B)}                        & 36.25             & 32.50                      & 3.7                                                                           \\ \hline
\end{tabular}
\end{table}

\subsection{Results}

\subsubsection{Performance and Efficiency Comparison}

Table \ref{tab:results-overall} consolidates the results, comparing our complete pipeline against the baselines. The findings highlight the effectiveness of our combined approach. Our final model (Prune + SFT + Quant) achieves a performance score of 54.25 while requiring only 3.9 GB of VRAM. This is a dramatic improvement over a naive 'Prune + Quant' approach, which scores an unusable 25.00. While performance is lower than the 13.4 GB original model, our method successfully reduces VRAM usage by 71\%, achieving a strong balance between performance and efficiency.

\begin{table}[h]
    \caption{Overall Results}
    \label{tab:results-overall}
\begin{tabular}{cccccc}
\hline
\textbf{Method}     & \textbf{\begin{tabular}[c]{@{}c@{}}Number of \\ Parameters(B)\end{tabular}} & \textbf{\begin{tabular}[c]{@{}c@{}}Bit\\ width\end{tabular}} & \textbf{\begin{tabular}[c]{@{}c@{}}Performance\\ Score\end{tabular}}  & \textbf{\begin{tabular}[c]{@{}c@{}}Latency \\ (ms/token)\end{tabular}} & \textbf{VRAM (GB)} \\ \hline
Original LLaVA      & 7.06                                                                        & 16                                                           & 80.12                  & 154                                                                    & 13.4               \\
Pruning Only        & 5.85                                                                        & 16                                                           & 62.50                  & 148                                                                    & 11.2               \\
Pruning and SFT     & 5.85                                                                        & 16                                                           & 74.25                  & 146                                                                    & 11.2               \\
Quantization Only   & 7.06                                                                        & 4                                                            & 64.72                  & 146                                                                    & 4.5                \\
Prune + Quant       & 5.85                                                                        & 4                                                            & 25.00                  & 122                                                                    & 3.9                \\
Prune + SFT + Quant & 5.85                                                                        & 4                                                            & 54.25                  & 122                                                                    & 3.9                \\ \hline
\end{tabular}
\end{table}
\subsubsection{Comparison}

To isolate the benefits of our specific component choices, we performed an ablation study detailed in Table~\ref{tab:ablation}. The results validate our methodology: replacing our activation-aware quantization (AWQ) with a standard Bitsandbytes technique causes a 7-point performance drop. Furthermore, using a simpler magnitude-based pruning baseline leads to a 4-point drop compared to our method. Attempting to use a standard round-to-neighbor quantization method resulted in unintelligible output, underscoring the necessity of our pipeline's advanced techniques.

\begin{table}[h]
    \centering
    \caption{Comparison Results}
    \label{tab:ablation}
    \begin{tabular}{lcc}
        \toprule
        \textbf{Setting} & \textbf{\begin{tabular}[c]{@{}c@{}}Performance\\ Score\end{tabular}}  & \textbf{Notes} \\
        \midrule
        Full pipeline (ours)                  & \textbf{54.25}  & Activation-aware quant + pruning + sft \\
        Bitsandbytes quantization                & 47           & $-$7\% accuracy drop \\
        Pruning Based on Magnitude Baseline      & 50           &  $-$4\% accuracy drop \\
        Round to neighbour Quantization          & 25           & unintelligible output \\
        \bottomrule
    \end{tabular}
\end{table}

\subsection{Case Studies}

\begin{figure}
    \centering
    \includegraphics[width=.8\linewidth]{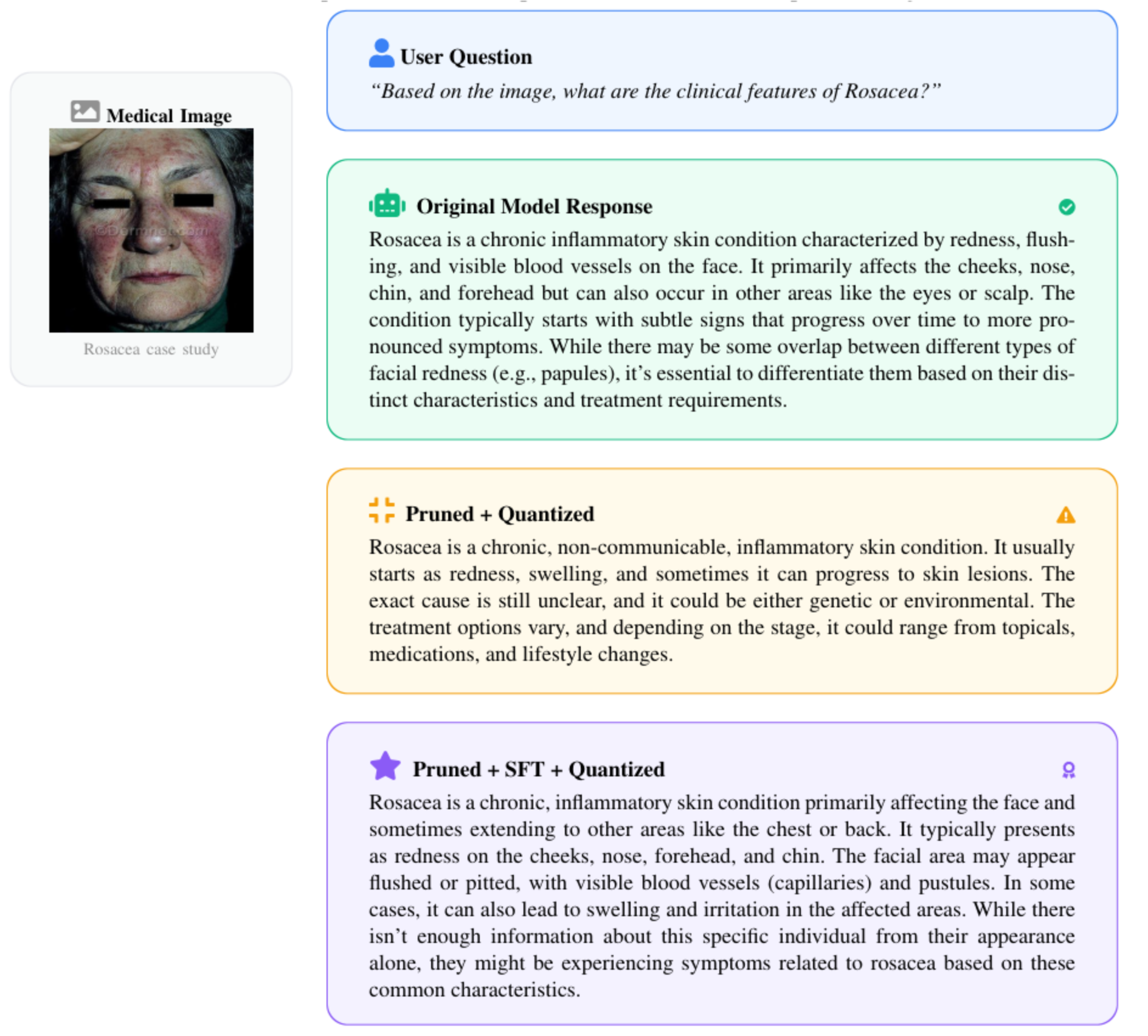}
    \caption{Comparison of model responses across different compression stages}
    \label{fig:model_comparison_dialog}
\end{figure}

Figure~\ref{fig:model_comparison_dialog} provides a qualitative comparison of model responses at different compression stages. The Original Model gives a detailed, accurate clinical description. After naive pruning and quantization (Pruned + Quantized), the model's response becomes generic and less clinically useful. However, the output from our full pipeline (Pruned + SFT + Quantized) recovers significant diagnostic detail, correctly identifying key characteristics of the condition. This case study demonstrates that our method preserves essential domain-specific knowledge even after significant compression.

\section{Discussion}\label{sec6}

This paper introduces an efficient and unified compression framework for deploying multimodal large language models (MLLMs) in memory-constrained medical environments, particularly for dermatological visual question answering (VQA). By jointly applying structural pruning and activation-aware post-training quantization, we significantly reduce the model's computational footprint while preserving its domain-specific performance. Our method demonstrates a 70\% reduction in VRAM usage and achieves a 4.2× inference speedup on NVIDIA T4 GPUs. Compared to isolated pruning or quantization strategies, our approach retains 4\% higher performance at the same compression ratio, validating the effectiveness of combining compression methods in a task-aware manner.

Despite these gains, several limitations remain. Our pruning strategy is limited to coarse-grained depth pruning of Transformer blocks, and activation-aware quantization is uniformly applied across layers. In future work, we plan to explore mixed-precision quantization, assigning higher precision to task-critical layers and lower precision elsewhere. Additionally, we aim to implement fine-grained pruning strategies—beyond entire Transformer blocks—to further optimize model structure. Finally, we intend to develop layer-wise optimization techniques that jointly determine the optimal mix of pruning and quantization per layer, enabling more adaptive and intelligent compression tailored to specific downstream medical tasks.

\backmatter












\newpage
\bibliography{sn-bibliography}


\begin{thebibliography}{16}
\ifx \bisbn   \undefined \def \bisbn  #1{ISBN #1}\fi
\ifx \binits  \undefined \def \binits#1{#1}\fi
\ifx \bauthor  \undefined \def \bauthor#1{#1}\fi
\ifx \batitle  \undefined \def \batitle#1{#1}\fi
\ifx \bjtitle  \undefined \def \bjtitle#1{#1}\fi
\ifx \bvolume  \undefined \def \bvolume#1{\textbf{#1}}\fi
\ifx \byear  \undefined \def \byear#1{#1}\fi
\ifx \bissue  \undefined \def \bissue#1{#1}\fi
\ifx \bfpage  \undefined \def \bfpage#1{#1}\fi
\ifx \blpage  \undefined \def \blpage #1{#1}\fi
\ifx \burl  \undefined \def \burl#1{\textsf{#1}}\fi
\ifx \doiurl  \undefined \def \doiurl#1{\url{https://doi.org/#1}}\fi
\ifx \betal  \undefined \def \betal{\textit{et al.}}\fi
\ifx \binstitute  \undefined \def \binstitute#1{#1}\fi
\ifx \binstitutionaled  \undefined \def \binstitutionaled#1{#1}\fi
\ifx \bctitle  \undefined \def \bctitle#1{#1}\fi
\ifx \beditor  \undefined \def \beditor#1{#1}\fi
\ifx \bpublisher  \undefined \def \bpublisher#1{#1}\fi
\ifx \bbtitle  \undefined \def \bbtitle#1{#1}\fi
\ifx \bedition  \undefined \def \bedition#1{#1}\fi
\ifx \bseriesno  \undefined \def \bseriesno#1{#1}\fi
\ifx \blocation  \undefined \def \blocation#1{#1}\fi
\ifx \bsertitle  \undefined \def \bsertitle#1{#1}\fi
\ifx \bsnm \undefined \def \bsnm#1{#1}\fi
\ifx \bsuffix \undefined \def \bsuffix#1{#1}\fi
\ifx \bparticle \undefined \def \bparticle#1{#1}\fi
\ifx \barticle \undefined \def \barticle#1{#1}\fi
\bibcommenthead
\ifx \bconfdate \undefined \def \bconfdate #1{#1}\fi
\ifx \botherref \undefined \def \botherref #1{#1}\fi
\ifx \url \undefined \def \url#1{\textsf{#1}}\fi
\ifx \bchapter \undefined \def \bchapter#1{#1}\fi
\ifx \bbook \undefined \def \bbook#1{#1}\fi
\ifx \bcomment \undefined \def \bcomment#1{#1}\fi
\ifx \oauthor \undefined \def \oauthor#1{#1}\fi
\ifx \citeauthoryear \undefined \def \citeauthoryear#1{#1}\fi
\ifx \endbibitem  \undefined \def \endbibitem {}\fi
\ifx \bconflocation  \undefined \def \bconflocation#1{#1}\fi
\ifx \arxivurl  \undefined \def \arxivurl#1{\textsf{#1}}\fi
\csname PreBibitemsHook\endcsname

\bibitem[\protect\citeauthoryear{Dettmers et~al.}{2022}]{dettmers2022llmint88bitmatrixmultiplication}
\begin{botherref}
\oauthor{\bsnm{Dettmers}, \binits{T.}},
\oauthor{\bsnm{Lewis}, \binits{M.}},
\oauthor{\bsnm{Belkada}, \binits{Y.}},
\oauthor{\bsnm{Zettlemoyer}, \binits{L.}}:
LLM.int8(): 8-bit Matrix Multiplication for Transformers at Scale
(2022).
\url{https://arxiv.org/abs/2208.07339}
\end{botherref}
\endbibitem

\bibitem[\protect\citeauthoryear{Badri and Shaji}{2023}]{badri2023hqq}
\begin{botherref}
\oauthor{\bsnm{Badri}, \binits{H.}},
\oauthor{\bsnm{Shaji}, \binits{A.}}:
Half-Quadratic Quantization of Large Machine Learning Models
(2023).
\url{https://mobiusml.github.io/hqq_blog/}
\end{botherref}
\endbibitem

\bibitem[\protect\citeauthoryear{Liu et~al.}{2023}]{Liu2023VisualIT}
\begin{botherref}
\oauthor{\bsnm{Liu}, \binits{H.}},
\oauthor{\bsnm{Li}, \binits{C.}},
\oauthor{\bsnm{Wu}, \binits{Q.}},
\oauthor{\bsnm{Lee}, \binits{Y.J.}}:
Visual instruction tuning.
ArXiv
\textbf{abs/2304.08485}
(2023)
\end{botherref}
\endbibitem

\bibitem[\protect\citeauthoryear{Wu et~al.}{2023}]{Wu2023MultimodalLL}
\begin{botherref}
\oauthor{\bsnm{Wu}, \binits{J.}},
\oauthor{\bsnm{Gan}, \binits{W.}},
\oauthor{\bsnm{Chen}, \binits{Z.}},
\oauthor{\bsnm{Wan}, \binits{S.}},
\oauthor{\bsnm{Yu}, \binits{P.S.}}:
Multimodal large language models: A survey.
2023 IEEE International Conference on Big Data (BigData),
2247--2256
(2023)
\end{botherref}
\endbibitem

\bibitem[\protect\citeauthoryear{Santomauro et~al.}{2023}]{Santomauro2023EnhancingMI}
\begin{bchapter}
\bauthor{\bsnm{Santomauro}, \binits{A.}},
\bauthor{\bsnm{Portinale}, \binits{L.}},
\bauthor{\bsnm{Leonardi}, \binits{G.}}:
\bctitle{Enhancing medical image report generation through standard language models: Leveraging the power of llms in healthcare}.
In: \bbtitle{HC@AIxIA}
(\byear{2023}).
\burl{https://api.semanticscholar.org/CorpusID:266211540}
\end{bchapter}
\endbibitem

\bibitem[\protect\citeauthoryear{Kim et~al.}{2024}]{kim2024shortened}
\begin{botherref}
\oauthor{\bsnm{Kim}, \binits{B.-K.}},
\oauthor{\bsnm{Kim}, \binits{G.}},
\oauthor{\bsnm{Kim}, \binits{T.-H.}},
\oauthor{\bsnm{Castells}, \binits{T.}},
\oauthor{\bsnm{Choi}, \binits{S.}},
\oauthor{\bsnm{Shin}, \binits{J.}},
\oauthor{\bsnm{Song}, \binits{H.-K.}}:
Shortened llama: A simple depth pruning for large language models.
arXiv preprint arXiv:2402.02834
\textbf{11}
(2024)
\end{botherref}
\endbibitem

\bibitem[\protect\citeauthoryear{Han et~al.}{2015}]{han2015learning}
\begin{botherref}
\oauthor{\bsnm{Han}, \binits{S.}},
\oauthor{\bsnm{Pool}, \binits{J.}},
\oauthor{\bsnm{Tran}, \binits{J.}},
\oauthor{\bsnm{Dally}, \binits{W.}}:
Learning both weights and connections for efficient neural network.
Advances in neural information processing systems
\textbf{28}
(2015)
\end{botherref}
\endbibitem

\bibitem[\protect\citeauthoryear{Men et~al.}{2024}]{men2024shortgptlayerslargelanguage}
\begin{botherref}
\oauthor{\bsnm{Men}, \binits{X.}},
\oauthor{\bsnm{Xu}, \binits{M.}},
\oauthor{\bsnm{Zhang}, \binits{Q.}},
\oauthor{\bsnm{Wang}, \binits{B.}},
\oauthor{\bsnm{Lin}, \binits{H.}},
\oauthor{\bsnm{Lu}, \binits{Y.}},
\oauthor{\bsnm{Han}, \binits{X.}},
\oauthor{\bsnm{Chen}, \binits{W.}}:
ShortGPT: Layers in Large Language Models are More Redundant Than You Expect
(2024).
\url{https://arxiv.org/abs/2403.03853}
\end{botherref}
\endbibitem

\bibitem[\protect\citeauthoryear{Liu et~al.}{2023}]{liu2023llm}
\begin{botherref}
\oauthor{\bsnm{Liu}, \binits{Z.}},
\oauthor{\bsnm{Oguz}, \binits{B.}},
\oauthor{\bsnm{Zhao}, \binits{C.}},
\oauthor{\bsnm{Chang}, \binits{E.}},
\oauthor{\bsnm{Stock}, \binits{P.}},
\oauthor{\bsnm{Mehdad}, \binits{Y.}},
\oauthor{\bsnm{Shi}, \binits{Y.}},
\oauthor{\bsnm{Krishnamoorthi}, \binits{R.}},
\oauthor{\bsnm{Chandra}, \binits{V.}}:
Llm-qat: Data-free quantization aware training for large language models.
arXiv preprint arXiv:2305.17888
(2023)
\end{botherref}
\endbibitem

\bibitem[\protect\citeauthoryear{Dettmers et~al.}{2023}]{dettmers2023qlora}
\begin{barticle}
\bauthor{\bsnm{Dettmers}, \binits{T.}},
\bauthor{\bsnm{Pagnoni}, \binits{A.}},
\bauthor{\bsnm{Holtzman}, \binits{A.}},
\bauthor{\bsnm{Zettlemoyer}, \binits{L.}}:
\batitle{Qlora: Efficient finetuning of quantized llms}.
\bjtitle{Advances in neural information processing systems}
\bvolume{36},
\bfpage{10088}--\blpage{10115}
(\byear{2023})
\end{barticle}
\endbibitem

\bibitem[\protect\citeauthoryear{Jeon et~al.}{2024}]{jeon2024l4q}
\begin{botherref}
\oauthor{\bsnm{Jeon}, \binits{H.}},
\oauthor{\bsnm{Kim}, \binits{Y.}},
\oauthor{\bsnm{Kim}, \binits{J.-j.}}:
L4q: Parameter efficient quantization-aware training on large language models via lora-wise lsq.
arXiv e-prints,
2402
(2024)
\end{botherref}
\endbibitem

\bibitem[\protect\citeauthoryear{Lin et~al.}{2024}]{lin2024awq}
\begin{barticle}
\bauthor{\bsnm{Lin}, \binits{J.}},
\bauthor{\bsnm{Tang}, \binits{J.}},
\bauthor{\bsnm{Tang}, \binits{H.}},
\bauthor{\bsnm{Yang}, \binits{S.}},
\bauthor{\bsnm{Chen}, \binits{W.-M.}},
\bauthor{\bsnm{Wang}, \binits{W.-C.}},
\bauthor{\bsnm{Xiao}, \binits{G.}},
\bauthor{\bsnm{Dang}, \binits{X.}},
\bauthor{\bsnm{Gan}, \binits{C.}},
\bauthor{\bsnm{Han}, \binits{S.}}:
\batitle{Awq: Activation-aware weight quantization for on-device llm compression and acceleration}.
\bjtitle{Proceedings of Machine Learning and Systems}
\bvolume{6},
\bfpage{87}--\blpage{100}
(\byear{2024})
\end{barticle}
\endbibitem

\bibitem[\protect\citeauthoryear{Kim}{2023}]{kim2023quantization}
\begin{barticle}
\bauthor{\bsnm{Kim}, \binits{J.}}:
\batitle{Quantization robust pruning with knowledge distillation}.
\bjtitle{IEEE Access}
\bvolume{11},
\bfpage{26419}--\blpage{26426}
(\byear{2023})
\end{barticle}
\endbibitem

\bibitem[\protect\citeauthoryear{Xu et~al.}{2021}]{xu2021accelerating}
\begin{barticle}
\bauthor{\bsnm{Xu}, \binits{W.}},
\bauthor{\bsnm{Fang}, \binits{W.}},
\bauthor{\bsnm{Ding}, \binits{Y.}},
\bauthor{\bsnm{Zou}, \binits{M.}},
\bauthor{\bsnm{Xiong}, \binits{N.}}:
\batitle{Accelerating federated learning for iot in big data analytics with pruning, quantization and selective updating}.
\bjtitle{IEEE Access}
\bvolume{9},
\bfpage{38457}--\blpage{38466}
(\byear{2021})
\end{barticle}
\endbibitem

\bibitem[\protect\citeauthoryear{Ma et~al.}{2023}]{ma2023llm}
\begin{barticle}
\bauthor{\bsnm{Ma}, \binits{X.}},
\bauthor{\bsnm{Fang}, \binits{G.}},
\bauthor{\bsnm{Wang}, \binits{X.}}:
\batitle{Llm-pruner: On the structural pruning of large language models}.
\bjtitle{Advances in neural information processing systems}
\bvolume{36},
\bfpage{21702}--\blpage{21720}
(\byear{2023})
\end{barticle}
\endbibitem

\bibitem[\protect\citeauthoryear{Wang et~al.}{2020}]{wang2020minilmdeepselfattentiondistillation}
\begin{botherref}
\oauthor{\bsnm{Wang}, \binits{W.}},
\oauthor{\bsnm{Wei}, \binits{F.}},
\oauthor{\bsnm{Dong}, \binits{L.}},
\oauthor{\bsnm{Bao}, \binits{H.}},
\oauthor{\bsnm{Yang}, \binits{N.}},
\oauthor{\bsnm{Zhou}, \binits{M.}}:
MiniLM: Deep Self-Attention Distillation for Task-Agnostic Compression of Pre-Trained Transformers
(2020).
\url{https://arxiv.org/abs/2002.10957}
\end{botherref}
\endbibitem

\end{thebibliography}

\begin{appendices}

\newpage
\section{LLM as a judge}\label{secA1}

\begin{figure}[h]
    \centering
    \includegraphics[width=0.75\linewidth]{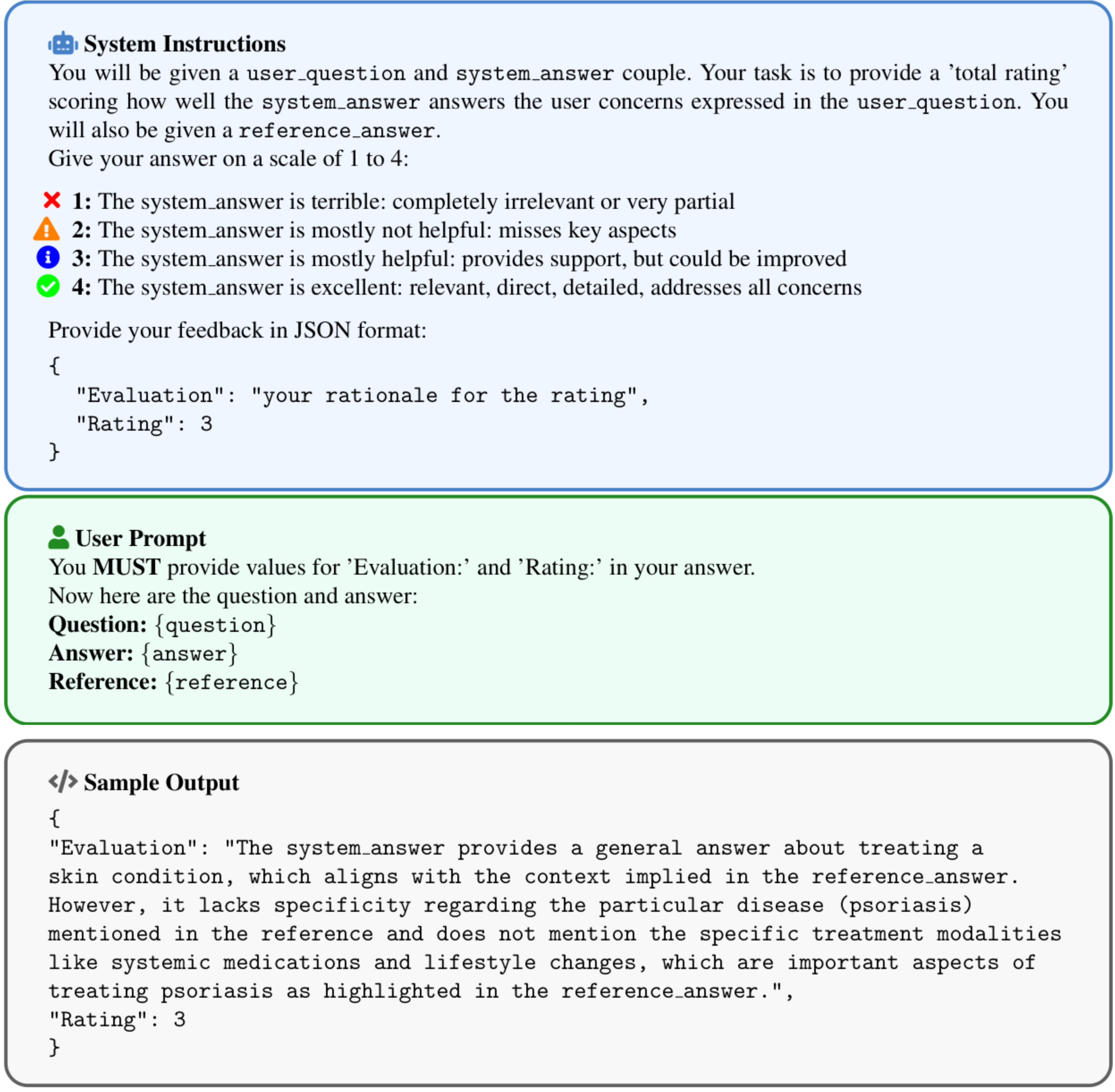}
    \caption{Evaluation Instruction Dialog}
    \label{fig:evaluation_table}
\end{figure}

We prepared 39 test images and 3 questions for each of them. We use Qwen-2.5-32B as our judge LLM. The following prompt was used to evaluate the performance of compressed model by comparing its generation against ground truth. After computing the average results, they are scaled into percentage-based scores. The prompts are detailed in Figure~\ref{fig:evaluation_table}

\newpage

\section{Result of Iterative Pruning}\label{secA2}
\begin{figure}[h]
    \centering
    \includegraphics[width=0.75\linewidth]{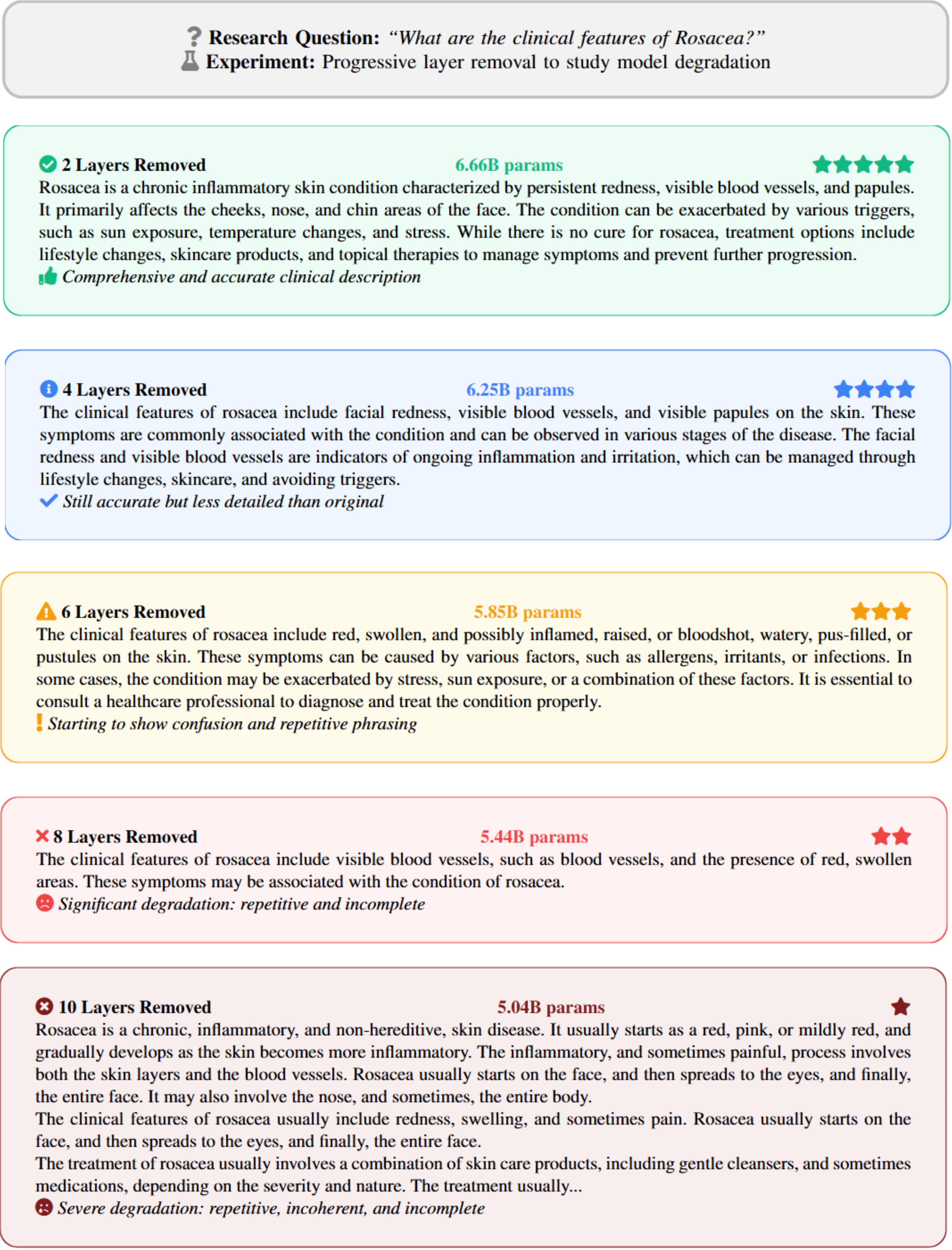}
    \caption{Model Output for Iterative Pruning - Performance Degradation Analysis}
    \label{fig:result_table}
\end{figure}

Figure~\ref{fig:result_table} shows sample output after iteratively pruning layers from the model. It can be seen that the output is deteriorating after pruning 6 layers. After pruning 10 layers, the model essentially looses the ability to generate the end of sequence token. 
\definecolor{excellent}{RGB}{16,185,129}
\definecolor{good}{RGB}{59,130,246}
\definecolor{fair}{RGB}{245,158,11}
\definecolor{poor}{RGB}{239,68,68}
\definecolor{terrible}{RGB}{127,29,29}

\definecolor{lightexcellent}{RGB}{236,253,245}
\definecolor{lightgood}{RGB}{239,246,255}
\definecolor{lightfair}{RGB}{255,251,235}
\definecolor{lightpoor}{RGB}{254,242,242}
\definecolor{lightterrible}{RGB}{248,238,238}

\end{appendices}

\end{document}